\documentclass[letterpaper]{article}
\usepackage{aaai2027}
\usepackage[hyphens]{url}  
\usepackage{graphicx} 
\usepackage{natbib}  
\usepackage{caption} 
\usepackage{algorithm}
\usepackage{algorithmic}

\usepackage{newfloat}
\usepackage{listings}
\DeclareCaptionStyle{ruled}{labelfont=normalfont,labelsep=colon,strut=off} 
\floatstyle{ruled}
\newfloat{listing}{tb}{lst}{}
\floatname{listing}{Listing}

\usepackage{booktabs}

\title{From Cellular Responses to Pharmacological Domains: Multimodal Zero-Shot Drug Representation Learning}

\author{
    Jintao Huang\textsuperscript{\rm 1},
    Lu Leng\textsuperscript{\rm 1}\corresponding,
    Ziyuan Yang\textsuperscript{\rm 2}\corresponding
}

\affiliations{
    \textsuperscript{\rm 1}Nanchang Hangkong University, Nanchang, Jiangxi, China\\
    \textsuperscript{\rm 2}Sichuan University, Chengdu, Sichuan, China\\
}

\usepackage{bibentry}
\usepackage{booktabs}
\usepackage{multirow}
\usepackage{graphicx}
\usepackage{booktabs}
\usepackage{multirow}
\usepackage[table]{xcolor}
\usepackage{amsmath}
\usepackage{amssymb}
\usepackage{tabularx}
\usepackage{array}
\usepackage{bm}
\definecolor{OursBlue}{RGB}{235,245,255}
\usepackage{booktabs}
\usepackage{tabularx}
\usepackage[table]{xcolor}
\definecolor{OursBlue}{RGB}{229,238,250}
\newcolumntype{Y}{>{\centering\arraybackslash}X}
\usepackage{subcaption}
\usepackage{algorithm}
\usepackage{algorithmic}
\usepackage{booktabs}
\usepackage{makecell}
\usepackage{graphicx}
\usepackage{newfloat}
\usepackage{listings}
\begin{document}

\maketitle

\begin{abstract}
Multimodal drug discovery enables drug representation learning beyond chemical structure by incorporating cellular responses such as gene expression and cell morphology. However, direct fusion and instance-level contrastive alignment may mix mechanism-related signals with modality-specific noise and incorrectly separate structurally dissimilar but biologically related compounds. This limitation can obscure transferable mechanism patterns required for predicting the properties of unseen compounds.
We introduce PMRD, a pharmacological response domain-guided framework for multimodal zero-shot drug property prediction. PMRD separates mechanism-consistent factors from modality-specific information and constructs a consensus response domain across three modalities. Mechanism candidate augmentation identifies locally stable factors, while retrieval-geometry attribution dynamically reweights the alignment and augmentation objectives according to whether their updates preserve inter-drug discriminability.This feedback suppresses training signals that conflict with mechanism-discriminative retrieval. PMRD further combines complementary representations through reliability-aware multiview retrieval.
Experiments on public datasets show improved zero-shot property prediction and more biologically coherent drug neighborhoods. Hard-negative analysis further indicates fewer conflicts between structurally dissimilar but response-related compounds. These results support PMRD as an effective framework for mechanism-aware multimodal drug representation learning.\footnote{The code will be released upon publication.}
\end{abstract}

\section{Introduction}
\label{sec:introduction}

In recent years, deep learning has significantly advanced
candidate compound screening, drug property prediction,
and mechanism-of-action analysis in computer-aided drug
discovery~\cite{rong2020grover}.
Drug property prediction aims to infer the biological effects
of compounds from limited observations. However, relying
solely on molecular structures is insufficient to fully
characterize the complex drug action process~\cite{subramanian2017l1000,bray2017cellpainting}.

\begin{figure}[t]
  \centering
  \includegraphics[
    width=\linewidth,
  ]{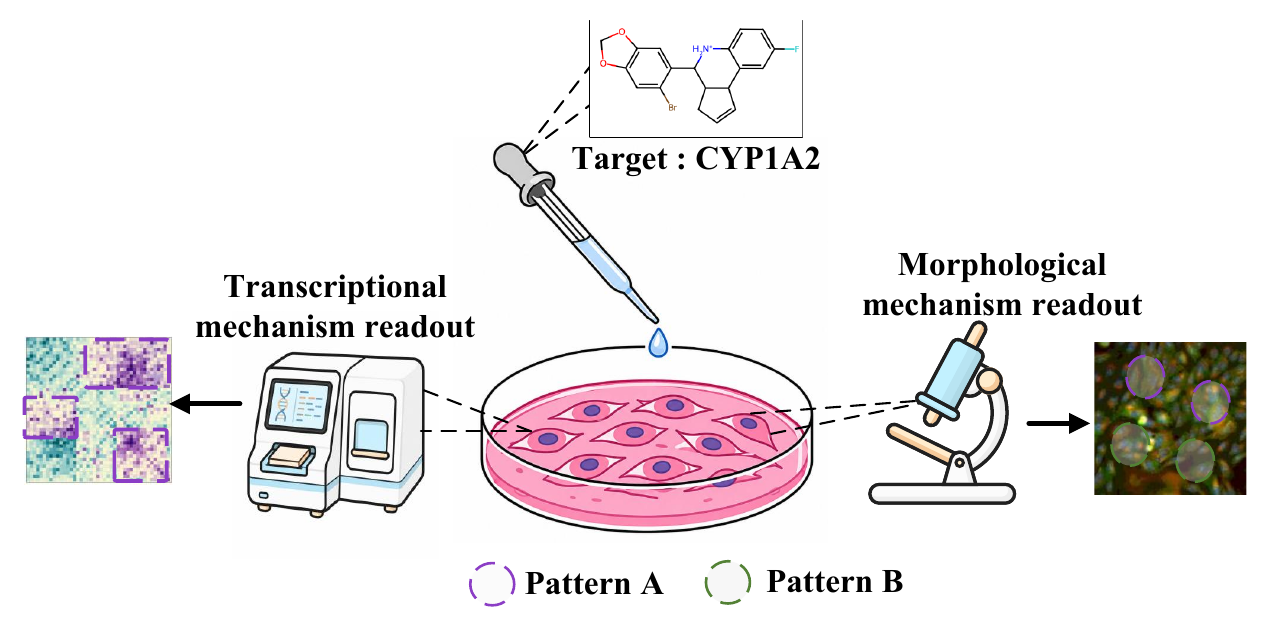}
 \caption{Conceptual overview of mechanism-aware drug profiling by integrating molecular structure with drug-induced transcriptional and morphological responses.}
  \label{fig:concept}
\end{figure}

To alleviate this information bottleneck, existing multimodal learning
approaches have begun to integrate molecular structures with biological
response data, such as gene expression profiles and cellular morphology.
As illustrated in Fig.~\ref{fig:concept}, these complementary observations
describe both the chemical identity of a compound and its induced cellular
responses, providing a broader basis for mechanism-aware drug representation
learning~\cite{wang2024infocore}.

Although multimodal information provides complementary perspectives for drug representation learning, its effective utilization remains limited by label scarcity and spurious
correlations~\cite{wu2018moleculenet,hu2020strategies,wang2024infocore}. High-quality drug--phenotype labels usually depend on expensive wet-lab experiments and are particularly scarce for
novel compounds, new targets, and unseen tasks, making it difficult for conventional supervised models to generalize to unseen scenarios~\cite{corsello2020prism}.
Meanwhile, molecular structures, gene expression profiles, and cellular morphology simultaneously contain pharmacologically relevant information, modality-specific
information, and experimental noise~\cite{subramanian2017l1000,bray2017cellpainting,mcquin2018cellprofiler,wang2024infocore}.
Directly fusing these observations may cause the model to mistake dataset biases or incidental co-occurrences for valid pharmacological signals, thereby weakening its transferability in zero-shot scenarios~\cite{du2023unimodal,wang2024infocore}.

Therefore, the key challenge in multimodal zero-shot drug representation learning is not merely to enhance statistical consistency across modalities, but to identify stable response factors that reflect drug action mechanisms from complex multi-source observations. This motivation is supported by prior findings that contrastive learning may rely on shortcut features~\cite{robinson2021shortcut}, high-throughput drug screens often contain batch-induced non-biological associations~\cite{wang2024infocore}, and multimodal models can suffer from modality dominance, where cleaner modalities suppress weaker but biologically informative views~\cite{du2023unimodal}. Moreover, weak molecule--phenotype alignment may make conventional negative sampling unreliable in phenotypic contrastive learning~\cite{rao2025miner}. Therefore, the model needs to determine which information consistently supports drug property discrimination across different modality views, sample neighborhoods, and task conditions, and which signals mainly arise from noise, bias, or local co-occurrence under specific data distributions. Based on this consideration, we focus on the following question:
\begin{quote}
\textit{How can stable drug response factors be identified from multi-source observations for reliable zero-shot property prediction?}
\end{quote}
To address this issue, we propose a multi-view zero-shot drug property prediction framework guided by a Pharmacological Mechanism Representation Domain~(PMRD). Since each modality contains both shared drug responses and source-specific variations, Cross-Modal Response Modeling~(CRM) decomposes its representation into a Mechanism-Consistent Representation~(MCR) and a Modality-Specific Representation~(MSR). MCR captures transferable pharmacological knowledge that remains consistent across heterogeneous modalities, whereas MSR preserves complementary modality-dependent characteristics. In this way, we can prevent modality-specific variations from interfering with mechanism-level alignment, thereby improving cross-modal knowledge transfer to unseen drug properties.

However, the above representations remain sensitive to variations, and to make it representation more robust and stable, we further propose a Stability-Guided Optimization~(SGO) module. Specifically, SGO consists of the Candidate Augmentation Mechanism~(CAM) and the Data Attribution Mechanism~(DAM) modules. CAM attempts to regularize the MCRs using locally perturbed candidates that preserve semantic and cross-modal consistency. DAM further assesses whether the updates preserve inter-drug discriminability and adaptively adjusts their weights accordingly.

Finally, since no labeled data are available in the zero-shot setting, our framework cannot train a conventional prediction head. We therefore introduce a Reliability-Aware Retrieval (RAR) process that performs property inference at test time by retrieving reliable reference samples from the learned representation space. Specifically, RAR retrieves the top-$K$ most similar drugs from the training set in each complementary representation view and aggregating their similarity-weighted predictions according to property-specific view reliability. This allows unseen drugs to exploit reliable property evidence without relying on a single fixed representation.Our main contributions are summarized as follows:
\begin{itemize}

    \item We propose a PMRD-guided framework for multimodal zero-shot drug
    property prediction, shifting the focus from fixed multimodal fusion to
    learning and utilizing stable drug-response evidence from heterogeneous
    observations.

    \item We develop CRM and SGO to construct cross-modally supported drug
    representations and refine them toward local stability and inter-drug
    discriminability, thereby reducing the influence of modality-specific bias
    and unreliable training signals.

    \item We introduce RAR to adaptively aggregate multiple retrieval views according to
    their property-specific reliability, enabling the model to exploit informative
    cross-view evidence without relying on a single representation.

\end{itemize}





\section{Related Work}
\label{sec:related_work}

\subsection{Molecular Representation Learning}

Molecular representation learning is a core problem in drug property prediction. Early methods use hand-crafted descriptors and fingerprints to encode local chemical substructures~\cite{rogers2010extended}. Deep models further represent molecules as atom--bond graphs and learn structural features through message passing or attention mechanisms~\cite{gilmer2017neural,yang2019dmpnn,xiong2020attentivefp}. Standard benchmarks such as MoleculeNet highlight the importance of scaffold split, label scarcity, and out-of-distribution generalization~\cite{wu2018moleculenet}.

To reduce reliance on task-specific labels, self-supervised molecular pre-training has been widely explored. Context prediction, node masking, edge prediction, and graph contrastive learning encourage transferable molecular representations before downstream fine-tuning~\cite{hu2020strategies,you2020graphcl,wang2022molclr}. Representative methods include GROVER, GraphLoG, and JOAO~\cite{rong2020grover,xu2021graphlog,you2021joao}. Other studies exploit 3D molecular geometry to capture spatial chemical information~\cite{zhou2023unimol,stark2022infomax}. However, these methods mainly rely on molecular structures and miss drug response factors observable only from phenotypic readouts.

\subsection{Multimodal Phenotypic Drug Representation Learning}

Phenotypic screening provides complementary biological evidence for modeling drug responses. LINCS L1000 measures transcriptional perturbations, while Cell Painting captures morphology changes induced by compounds~\cite{subramanian2017l1000,bray2017cellpainting,mcquin2018cellprofiler}. Large-scale resources such as JUMP Cell Painting further support phenotype-aware drug discovery~\cite{chandrasekaran2023jump}. These readouts have been used for mechanism-of-action analysis, drug repurposing, and phenotype-based retrieval~\cite{corsello2020prism,jang2021moa}.

Recent multimodal methods align molecular structures with phenotypic observations. Inspired by CLIP-style contrastive alignment~\cite{radford2021clip}, CLOOME aligns chemical structures with cellular images~\cite{sanchezfernandez2023cloome}, MIGA performs graph--image contrastive learning~\cite{zheng2024miga}, InfoCORE reduces batch-effect bias via conditional mutual information maximization~\cite{wang2024infocore}, and MINER calibrates negative sampling for incomplete multimodal observations~\cite{rao2025miner}. These methods validate the usefulness of phenotypic data, but mainly focus on cross-modal consistency or supervised downstream prediction, leaving stable mechanism-related factor selection under zero-shot settings less explored.

\subsection{Zero-shot and Robust Multimodal Prediction}

Zero-shot drug property prediction aims to infer the properties of unseen compounds without task-specific supervised fine-tuning by exploiting transferable representations learned from chemical structures and perturbation-derived phenotypes~\cite{sanchez2023cloome,liu2025learning}. In multimodal settings, this problem is complicated by the coexistence of pharmacologically relevant responses, modality-specific variations, batch effects, and experimental noise~\cite{zheng2024crossmodal,wang2024infocore}. Optimization may also exhibit modality dominance, where stronger or cleaner modalities suppress weaker but biologically informative views~\cite{du2023unimodal}. The central challenge is therefore to preserve transferable and complementary cross-modal evidence while limiting interference from source-specific variation, technical confounders, and dominant modalities.

\section{Proposed Method}

\begin{figure*}[!t]
  \centering
  \includegraphics[width=\textwidth]{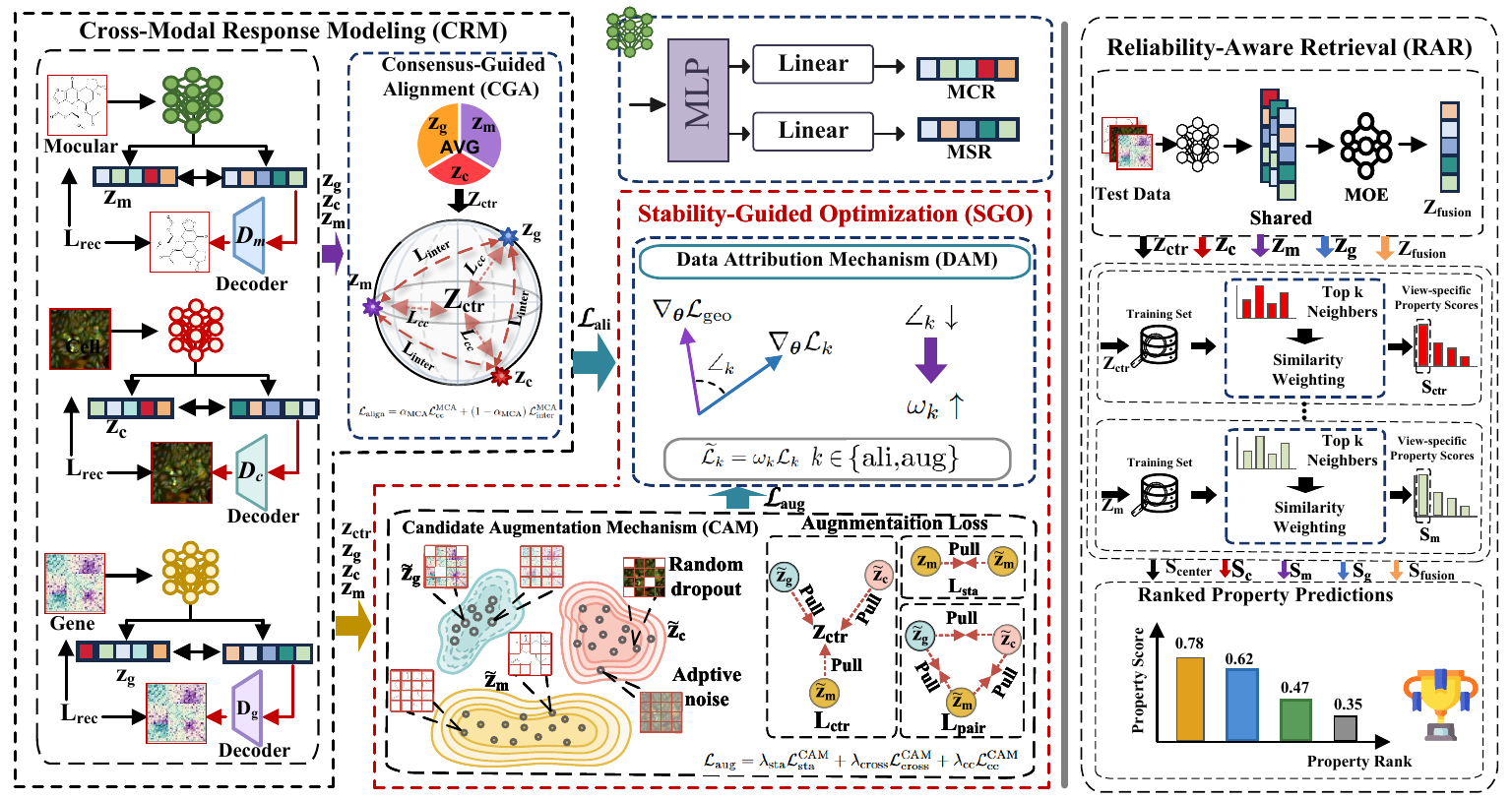}
  \caption{Overview of the proposed PMRD framework. PMRD disentangles multimodal drug observations into mechanism-consistent and modality-specific representations, constructs a cross-modal consensus mechanism domain, and uses attribution-guided loss reweighting to retain effective training signals. The learned representations are subsequently integrated through multiview retrieval for zero-shot drug property prediction.}
  \label{fig:kuangjia_overview}
\end{figure*}

\subsection{Overview}


To identify stable mechanistic evidence from heterogeneous drug observations,
we propose a PMRD-guided framework for multimodal zero-shot drug property
prediction, as illustrated in Fig.~\ref{fig:kuangjia_overview}. Given molecular
structure, gene expression, and cellular morphology, the framework proceeds
through CRM, SGO, and RAR.
CRM directs shared drug responses and source-specific variations into the MCR
and MSR, respectively, preventing modality bias from disrupting cross-modal
knowledge transfer. CGA then constructs a drug-specific PMRD consensus center
from the three MCRs and aligns them with this reference while preserving
inter-drug distinctions.
Because the shared responses identified by CRM may depend on incidental
features, SGO uses CAM to retain factors whose semantics and cross-modal
relations persist under local perturbations. DAM further adjusts the weights
of CGA and CAM according to whether their updates preserve inter-drug
discriminability.
During zero-shot prediction, task-specific fine-tuning is unavailable and view
reliability varies across properties.RAR adaptively aggregates multi-view predictions based on the reliability of different attributes from each view, avoiding
dependence on a single fixed representation.

\subsection{Cross-Modal Response Modeling}
\label{sec:crm}

Drug observations contain both cross-modal response patterns and
modality-specific variations. Direct alignment may therefore introduce
source-dependent bias into the space. CRM addresses this issue by
separating the two factors and consolidating response information supported
across modalities.

For drug $i$ in modality $m$, two projection heads produce MCR
$\mathbf{z}_i^m$ and MSR $\mathbf{u}_i^m$, where
$m\in\mathcal{M}=\{\mathrm{mol},\mathrm{gene},\mathrm{cell}\}$.
MCR participates in CGA and retains transferable response information,
whereas MSR reconstructs only its own modality and preserves source-specific
variation. CRM further reduces redundancy between the two branches as follows:
\begin{equation}
\mathcal{L}_{\mathrm{sep}}
=
\operatorname{Avg}_{i,m}
\left[
\operatorname{sim}\!\left(
\mathbf{z}_i^m,\mathbf{u}_i^m
\right)^2
\right].
\label{eq:crm_separation}
\end{equation}

Minimizing Eq.~\eqref{eq:crm_separation} reduces information overlap between
MCR and MSR, while reconstruction prevents trivial separation through
information loss.

A single MCR may still retain correlations supported by only one modality.
CGA therefore averages the three normalized MCRs of each drug and normalizes
the result to construct a drug-specific PMRD center. This center strengthens
response directions shared across modalities and suppresses unsupported
variations.

CGA combines center alignment and inter-modal alignment as follows:
\begin{equation}
\mathcal{L}_{\mathrm{ali}}
=
\alpha_{\mathrm{CGA}}
\mathcal{L}_{\mathrm{cc}}^{\mathrm{CGA}}
+
\left(1-\alpha_{\mathrm{CGA}}\right)
\mathcal{L}_{\mathrm{inter}}^{\mathrm{CGA}}.
\label{eq:cga_alignment}
\end{equation}

Here, $\mathcal{L}_{\mathrm{cc}}^{\mathrm{CGA}}$ aligns each MCR with the
PMRD center of the same drug, whereas
$\mathcal{L}_{\mathrm{inter}}^{\mathrm{CGA}}$ directly aligns its three MCRs
to prevent center averaging from masking cross-modal disagreement. Other drugs
in the mini-batch serve as negatives to preserve inter-drug discriminability,
and $\alpha_{\mathrm{CGA}}$ balances the two objectives.
\subsection{Stability-Guided Optimization}
\label{sec:sgo}

CGA establishes cross-modal response consistency at the current representation points,
but this agreement may rely on incidental features and disappear under mild
perturbations. Moreover, CGA and CAM may not be equally beneficial in every
mini-batch, and excessive alignment or augmentation can weaken inter-drug
discriminability. SGO addresses these issues through CAM and DAM: CAM retains
locally stable response factors, while DAM selects optimization signals that
preserve drug-level distinctions.

CAM perturbs each MCR using random feature dropout and scale-adaptive Gaussian
noise, and retains candidates that preserve semantics, cross-modal relations,
PMRD-center consistency, and the local distribution. Samples without valid
candidates contribute no augmentation loss. The selected candidates regularize
the original MCRs as follows:
\begin{equation}
\mathcal{L}_{\mathrm{aug}}
=
\lambda_{\mathrm{sta}}
\mathcal{L}_{\mathrm{sta}}^{\mathrm{CAM}}
+
\lambda_{\mathrm{cross}}
\mathcal{L}_{\mathrm{cross}}^{\mathrm{CAM}}
+
\lambda_{\mathrm{cc}}
\mathcal{L}_{\mathrm{cc}}^{\mathrm{CAM}},
\label{eq:cam_augmentation}
\end{equation}
where $\mathcal{L}_{\mathrm{sta}}^{\mathrm{CAM}}$ preserves candidate
semantics and local structure,
$\mathcal{L}_{\mathrm{cross}}^{\mathrm{CAM}}$ maintains cross-modal
consistency, and $\mathcal{L}_{\mathrm{cc}}^{\mathrm{CAM}}$ aligns the
candidates with the corresponding PMRD center. The candidates are used only
to regularize local MCR neighborhoods and do not replace the original MCRs or
participate in retrieval.

Local stability alone does not guarantee separation between different drugs.
DAM therefore defines an attribution reference from the multi-view retrieval
geometry as follows:
\begin{equation}
\mathcal{L}_{\mathrm{geo}}
=
\operatorname{Avg}_{v\in\mathcal{V},\,i\neq j}
\left[
\operatorname{sim}\!\left(
\mathbf{h}_i^v,\mathbf{h}_j^v
\right)^2
\right],
\label{eq:dam_geometry}
\end{equation}
where
$\mathcal{V}=
\{\mathrm{mol},\mathrm{gene},\mathrm{cell},\mathrm{ctr},\mathrm{fus}\}$
contains the three MCRs, the PMRD center, and the fusion representation.
Minimizing Eq.~\eqref{eq:dam_geometry} suppresses excessive similarity between
different drugs, so its descent direction indicates whether the updates
induced by CGA and CAM preserve inter-drug discriminability.

DAM converts this directional support into dynamic loss weights as follows:
\begin{equation}
\widetilde{\mathcal{L}}_k
=
\omega_k\mathcal{L}_k.
\label{eq:dam_reweighting}
\end{equation}

Here, $k\in\{\mathrm{ali},\mathrm{aug}\}$ denotes the CGA and CAM objectives,
respectively. Objectives aligned with the discriminative direction receive
larger weights, whereas conflicting signals are suppressed.
$\mathcal{L}_{\mathrm{geo}}$ serves only as the attribution reference and is
not added to the training loss; the reconstruction loss retains a fixed
weight. SGO therefore preserves response factors that are both locally stable
and discriminative across drugs.
\subsection{Reliability-Aware Retrieval}
\label{sec:rar}

Unseen query drugs have no property labels, and the zero-shot protocol
precludes task-specific fine-tuning. RAR therefore infers their properties by
retrieving similar labeled drugs from the training set. Because a retrieval
view may be reliable for one property but less informative for another, RAR
evaluates five views independently: the three MCRs, the PMRD center, and the
MoE fusion representation.

In each view, RAR retrieves the top-$K$ training drugs by cosine similarity
and computes a similarity-weighted prediction from their valid labels. Missing
labels are excluded; if none of the retrieved drugs is labeled for a property,
RAR uses its positive-label proportion in the training set as the fallback
score.

RAR then aggregates the view-specific predictions according to their
property-specific reliability as follows:
\begin{equation}
\hat{y}_{t}(q)
=
\sum_{v\in\mathcal{V}}
\rho_{v,t}\hat{y}_{v,t}(q),
\label{eq:rar_aggregation}
\end{equation}
where $\hat{y}_{v,t}(q)$ is the prediction from view $v$ for property $t$ of
query drug $q$, and $\rho_{v,t}$ is its normalized reliability weight.
Reliability is estimated from prediction confidence, agreement among views,
neighborhood overlap, label coverage, and the attribution prior. The learned
representations determine which training drugs are retrieved, while RAR
determines which views should be trusted for each property.
\section{Experiments}
\label{sec:experiments}
\begin{table*}[t]
\centering

\small
\setlength{\tabcolsep}{3.6pt}
\renewcommand{\arraystretch}{1.05}
\begin{tabular}{@{}llccccccc@{}}
\toprule
\multicolumn{2}{c}{Datasets} &
\multicolumn{4}{c}{ChEMBL2K~(AUROC $\uparrow$)} &
\multicolumn{3}{c}{Broad6K~(AUROC $\uparrow$)} \\
\cmidrule(lr){3-6}\cmidrule(lr){7-9}
\multicolumn{2}{c}{Metric} & Avg. & $>80\%$ & $>85\%$ & $>90\%$ &
Avg. & $>80\%$ & $>85\%$ \\
\midrule
\multirow{3}{*}{Supervised}
& Gene Expression & 56.1$\pm$1.1 & 5.1$\pm$1.4 & 3.4$\pm$1.3 & 3.4$\pm$1.3 & 56.9$\pm$1.4 & 1.9$\pm$1.7 & 1.9$\pm$1.7 \\
& Cell Morphology & 64.3$\pm$2.4 & 15.6$\pm$6.6 & 8.3$\pm$3.7 & 4.9$\pm$3.9 & 55.3$\pm$0.1 & 0.0$\pm$0.0 & 0.0$\pm$0.0 \\
& Morgan FP & \underline{76.8$\pm$2.2} & 48.8$\pm$3.9 & 34.6$\pm$6.3 & \underline{21.9$\pm$5.7} & 63.3$\pm$0.3 & 6.3$\pm$0.0 & \textbf{4.4$\pm$1.7} \\
\midrule
\multirow{5}{*}{GNN Pretraining}
& AttrMask & 73.3$\pm$1.4 & 46.3$\pm$3.4 & 30.3$\pm$2.4 & 14.6$\pm$1.7 & 59.8$\pm$0.2 & 3.1$\pm$0.0 & 3.1$\pm$0.0 \\
& EdgePred & 75.6$\pm$0.5 & 54.2$\pm$4.0 & 34.6$\pm$7.2 & 12.2$\pm$2.4 & 59.9$\pm$0.2 & 3.1$\pm$0.0 & 3.1$\pm$0.0 \\
& GROVER & 73.3$\pm$1.4 & 38.5$\pm$2.0 & 22.4$\pm$3.6 & 14.6$\pm$2.4 & 66.2$\pm$0.1 & \textbf{15.6$\pm$0.0} & \underline{3.8$\pm$1.4} \\
& GraphLoG & 73.5$\pm$0.7 & 41.9$\pm$2.0 & 29.3$\pm$3.4 & 15.6$\pm$2.8 & 62.9$\pm$0.4 & 4.4$\pm$1.7 & 0.0$\pm$0.0 \\
& JOAO & 75.1$\pm$1.0 & 47.8$\pm$5.1 & 33.7$\pm$2.0 & 19.0$\pm$3.2 & 67.3$\pm$0.4 & \underline{12.5$\pm$0.0} & \underline{3.8$\pm$1.4} \\
\midrule
\multirow{3}{*}{Multimodal Alignment}
& CLOOME & 66.7$\pm$1.8 & 26.8$\pm$4.6 & 16.1$\pm$3.7 & 10.7$\pm$5.1 & 61.7$\pm$0.4 & 3.1$\pm$0.0 & 3.1$\pm$0.0 \\
& MIGA & 75.6$\pm$1.6 & \textbf{55.1$\pm$1.3} & \underline{35.1$\pm$4.4} & 14.6$\pm$1.7 & 62.9$\pm$0.4 & 4.4$\pm$1.7 & 0.0$\pm$0.0 \\

\midrule
\multirow{3}{*}{Zero-shot Alignment}
& CLIP$^\dagger$ & 71.1$\pm$0.8 & 37.0$\pm$2.8 & 20.9$\pm$4.5 & 15.1$\pm$3.9 & \underline{67.4$\pm$0.3} & 9.3$\pm$1.9 & 3.1$\pm$0.0 \\
& CCL$^\dagger$ & 71.5$\pm$0.8 & 33.1$\pm$5.0 & 22.9$\pm$4.7 & 15.6$\pm$2.4 & 67.2$\pm$0.5 & 10.0$\pm$2.3 & 3.1$\pm$0.0 \\
& \cellcolor{OursBlue}\textbf{Ours}$^\dagger$
& \cellcolor{OursBlue}\textbf{77.8$\pm$0.3}
& \cellcolor{OursBlue}\underline{54.6$\pm$3.7}
& \cellcolor{OursBlue}\textbf{37.1$\pm$3.2}
& \cellcolor{OursBlue}\textbf{25.4$\pm$3.7}
& \cellcolor{OursBlue}\textbf{67.9$\pm$0.5}
& \cellcolor{OursBlue}11.2$\pm$3.6
& \cellcolor{OursBlue}\textbf{4.4$\pm$1.7} \\
\bottomrule
\end{tabular}
\caption{Molecular property prediction on ChEMBL2K and Broad6K. We report mean
AUROC and the proportions of valid tasks whose AUROC exceeds the indicated
thresholds. Best and second-best results are marked in \textbf{bold} and
\underline{underline}, respectively. $^\dagger$ denotes zero-shot methods that
do not use target-property labels for task-specific supervised fine-tuning.}
\label{tab:property_prediction}
\end{table*}
\begin{table}[t]
\centering

\footnotesize
\setlength{\tabcolsep}{2.5pt}
\renewcommand{\arraystretch}{1.08}

\begin{tabularx}{\columnwidth}{
@{}>{\raggedright\arraybackslash}Xcccc@{}
}
\toprule
\textbf{Method}
& \textbf{Avg.}
& $\mathbf{>80\%}$
& $\mathbf{>85\%}$
& $\mathbf{>90\%}$ \\
\midrule

InfoCORE~(CP)
& $73.8{\pm}2.0$
& $37.6{\pm}9.2$
& $26.3{\pm}4.7$
& $10.7{\pm}4.1$ \\

InfoCORE~(GE)
& $79.3{\pm}0.9$
& $62.4{\pm}2.8$
& $46.3{\pm}3.0$
& $30.3{\pm}2.2$ \\

MINER
& $81.2{\pm}1.0$
& $63.9{\pm}3.2$
& $50.2{\pm}4.2$
& $\mathbf{37.5{\pm}4.5}$ \\

\midrule

\rowcolor{OursBlue}
\textbf{Ours variant}
& $\mathbf{81.6{\pm}1.0}$
& $\mathbf{67.8{\pm}2.8}$
& $\mathbf{54.6{\pm}3.7}$
& $36.6{\pm}4.4$ \\

\bottomrule
\end{tabularx}
\caption{AUROC comparison on ChEMBL2K. Values are mean
$\pm$ standard deviation; the best result is shown in bold.}
\label{tab:chembl2k_label_comparison}
\end{table}
\begin{figure*}[t]
    \centering
    \includegraphics[width=\textwidth]{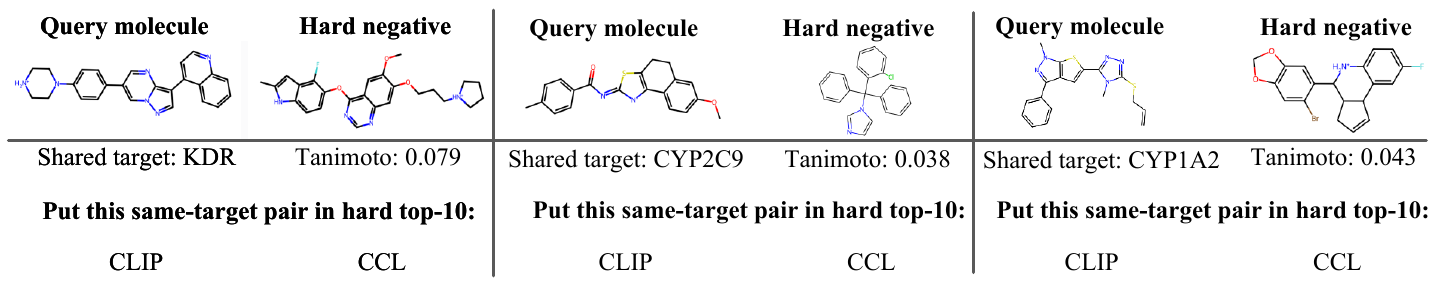}
    \caption{Representative molecules returned under the low-structure
    hard-negative retrieval protocol.}
    \label{fig:low_structure_hard_case}
\end{figure*}
\begin{figure}[t]
    \centering
    \begin{subfigure}[t]{\columnwidth}
        \centering
        \includegraphics[
            width=0.88\linewidth,
        ]{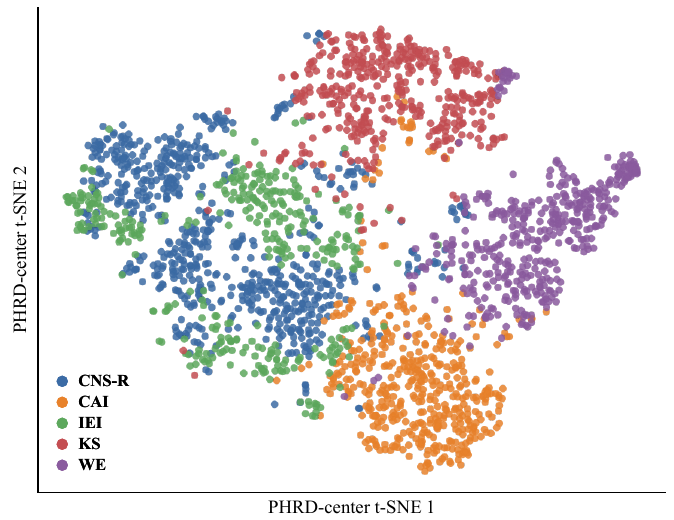}
        \caption{ChEMBL2K}
        \label{fig:tsne_chembl2k}
    \end{subfigure}

    \vspace{1pt}

    \begin{subfigure}[t]{\columnwidth}
        \centering
        \includegraphics[
            width=0.88\linewidth,
        ]{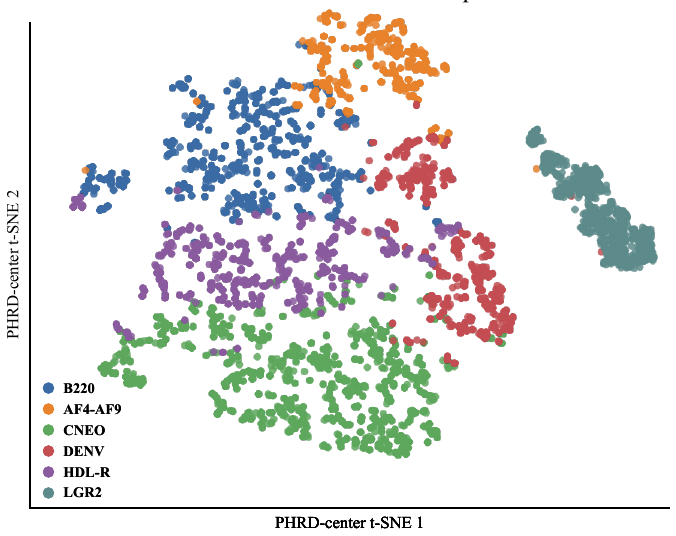}
        \caption{Broad6K}
        \label{fig:tsne_broad6k}
    \end{subfigure}

    \caption{t-SNE visualization of PMRD-center representations on ChEMBL2K
    and Broad6K. Colors denote cluster-level enriched biological or assay
    themes.}
    \label{fig:mechanism_enriched_tsne}
\end{figure}
\subsection{Experimental Setup}
\label{sec:experimental_setup}

\subsubsection{Datasets}
\label{sec:datasets_splits}
We conduct experiments on two multimodal drug property datasets, ChEMBL2K~\cite{gaulton2012chembl} and Broad6K~\cite{moshkov2023predicting}. ChEMBL2K contains 2,355 drug samples and 41 binary property prediction tasks, while Broad6K contains 6,567 drug samples and 32 binary property prediction tasks. Each sample includes three types of observations: molecular structure, gene expression, and cellular morphology. To evaluate predictive performance on unseen drugs, we adopt a scaffold split and partition each dataset into training, validation, and test sets with a ratio of $0.60/0.15/0.25$. ChEMBL2K is divided into 1,413 training, 353 validation, and 589 test samples, respectively. Broad6K is divided into 3,940 training, 985 validation, and 1,642 test samples, respectively.

\subsubsection{Evaluation Metrics}
\label{sec:evaluation_metrics}

We use the area under the receiver operating characteristic curve~(AUROC) as
the primary evaluation metric. In addition to mean AUROC across tasks, we
report the proportions of tasks achieving AUROC values above 0.80, 0.85, and,
where applicable, 0.90. Unless otherwise specified, results are reported as
the mean and standard deviation over five independent runs.

\subsubsection{Baseline}
\label{sec:baselines_protocol}

We compare our method with four groups of baseline methods. The supervised
unimodal group contains models based on gene expression, cellular morphology,
and Morgan fingerprints. The GNN-pretraining group includes AttrMask
, EdgePred \cite{hu2020strategies}, GROVER \cite{rong2020grover}, GraphLoG \cite{xu2021graphlog}, and JOAO \cite{you2021joao}. The multimodal-alignment group includes CLOOME \cite{sanchez2023cloome}, MIGA \cite{zheng2024crossmodal}, and InfoCORE \cite{wang2024infocore}. The zero-shot alignment group contains CLIP \cite{radford2021learning}, CCL\cite{ma2021conditional}, and PMRD~(Ours).



\subsubsection{Implementation Details}
\label{sec:implementation_details}

The molecular structure encoder is implemented as a three-layer Graph
Isomorphism Network~(GIN), whereas the gene-expression and cellular-morphology
encoders are implemented as five-layer deep neural networks. The hidden
dimension is 64. We train the model for 240 epochs with a batch size of 1024
and a learning rate of 0.005.

During zero-shot retrieval, we use the top-20 nearest neighbors and set the
retrieval temperature to 0.1. For PMRD alignment, the weights of reference
alignment and inter-modality alignment are 0.6 and 0.4, respectively. The
stable candidate augmentation module generates eight local candidates, and
its augmentation weight is 0.3. During inference, we retain five complementary
views: the fusion view, molecular MCR view, gene MCR view, cellular MCR view,
and PMRD view. Their predictions are aggregated using task-level weights based
on view reliability.
\subsection{Molecular Property Prediction}
\label{sec:property_prediction}

We evaluate PMRD under both strict zero-shot and label-informed settings. In
the strict zero-shot setting, target-property labels are excluded from
representation learning, and predictions are obtained solely through
reliability-aware retrieval over the five learned views.
Table~\ref{tab:property_prediction} compares PMRD with supervised unimodal,
GNN-pretraining, multimodal-alignment, and zero-shot baselines on ChEMBL2K and
Broad6K.

On ChEMBL2K, PMRD achieves the highest average AUROC of 77.8\%, exceeding the
strongest supervised unimodal baseline, Morgan FP, by 1.0 percentage point. It
also obtains the best proportions of tasks above the 0.85 and 0.90 thresholds,
reaching 37.1\% and 25.4\%, respectively. Although MIGA is slightly higher at
the 0.80 threshold, PMRD performs better at the stricter thresholds, indicating
that its gains extend beyond marginal improvements and benefit a larger number
of high-performing tasks.

On Broad6K, PMRD again achieves the highest average AUROC of 67.9\%,
outperforming CLIP and JOAO by 0.5 and 0.6 percentage points, respectively. It
also ties for the best proportion of tasks above 0.85. While PMRD does not lead
at the 0.80 threshold, its superior average performance suggests a more
balanced improvement across properties rather than gains concentrated in a
small subset of tasks. These results support the central motivation of PMRD: separating
modality-specific variation, retaining locally stable response factors, and
selecting reliable retrieval views can improve the transferability of
multimodal drug representations to unseen property tasks without
task-specific supervision.

\paragraph{Extension with Property-Label Supervision.}

To examine whether the learned mechanism representations remain useful when
property labels are available, we further introduce \textbf{Ours + Label
Loss}. This variant adds a masked prediction loss over observed training labels
while retaining the original PMRD objectives. It is evaluated separately
because it no longer follows the strict zero-shot protocol.

As shown in Table~\ref{tab:chembl2k_label_comparison}, the label-informed
variant increases the average AUROC from 77.8\% to 81.6\%. It achieves the best
average AUROC and the highest proportions of tasks above 0.80 and 0.85.
Compared with MINER, it improves these three metrics by 0.4, 3.9, and 4.4
percentage points, respectively, while remaining 0.9 points lower at the 0.90
threshold.

The larger gains at the 0.80 and 0.85 thresholds indicate that property
supervision improves performance across a broader range of tasks, rather than
only increasing the number of exceptionally high-scoring tasks. This result
also shows that the mechanism representations learned by PMRD are compatible
with direct supervision: they support zero-shot retrieval when labels are
unavailable and can be further refined when task labels are provided.
\subsection{Low-Structure Hard-Negative Analysis}
\label{sec:hard_negative_analysis}

Instance-level contrastive alignment may separate compounds solely because they
are structurally dissimilar, even when they act on related biological targets.
Such pairs constitute potentially conflicting negatives: pushing them apart
may disrupt pharmacologically meaningful neighborhoods. We examine this issue
by retaining compound pairs with Morgan Tanimoto similarity below 0.1 and
measuring the same-target rate among the top-$K$ non-self negative candidates.
A candidate is counted as conflicting when it shares at least one annotated
target with the query. Lower values therefore indicate that fewer
target-overlapping compounds are selected as negatives.

\begin{table}[t]
\centering

\small
\setlength{\tabcolsep}{4.2pt}
\renewcommand{\arraystretch}{1.12}

\begin{tabularx}{\columnwidth}{
@{}l *{3}{>{\centering\arraybackslash}X}@{}
}
\toprule
Method
& Top-10 $\downarrow$
& Top-20 $\downarrow$
& Top-30 $\downarrow$ \\
\midrule

CLIP          & 23.60\% & 23.63\% & 23.00\% \\
CCL           & 22.73\% & 22.34\% & 22.11\% \\
InfoCORE~(CP) & 22.15\% & 20.77\% & 20.82\% \\
MINER         & 27.75\% & 29.30\% & 28.75\% \\

\rowcolor{OursBlue}
\textbf{Ours}
& \textbf{10.69\%}
& \textbf{10.38\%}
& \textbf{10.83\%} \\

\bottomrule
\end{tabularx}
\caption{Same-target rate~(\%) among the Top-$K$ non-self candidates after
filtering structurally similar pairs by Morgan Tanimoto similarity. It measures
the percentage of candidate compounds sharing at least one annotated target
with the query; lower is better.}
\label{tab:low_structure_hard_retrieval}

\end{table}

As shown in Table~\ref{tab:low_structure_hard_retrieval}, PMRD maintains a
same-target rate of 10.38--10.83\% across the three retrieval depths, whereas
the baselines range from 20.77\% to 29.30\%. Compared with the lowest baseline
at each cutoff, InfoCORE~(CP), PMRD reduces the rate by 11.46, 10.39, and
9.99 percentage points at Top-10, Top-20, and Top-30, respectively. Under the
reported protocol, these results indicate that the learned retrieval space is
less likely to place structurally dissimilar but target-overlapping compounds
among negative candidates.

Figure~\ref{fig:low_structure_hard_case} presents representative Top-10 cases.
Although the displayed query--candidate pairs have Morgan Tanimoto similarities
below 0.1, several baseline-selected candidates share annotated targets such as
KDR, CYP2C9, and CYP1A2 with their queries. These examples illustrate how
structural dissimilarity can coexist with biological relatedness and why such
pairs may be inappropriate negatives. Because target annotations are incomplete
and target overlap captures only one aspect of pharmacological mechanism, this
analysis should be interpreted as evidence of reduced annotated target
conflicts rather than a complete validation of mechanism equivalence.
\subsection{Ablation Study}
\label{sec:ablation_study}

\begin{table}[t]
\centering

\small
\setlength{\tabcolsep}{4.2pt}
\renewcommand{\arraystretch}{1.12}

\begin{tabularx}{\columnwidth}{
@{}l *{3}{>{\centering\arraybackslash}X}@{}
}
\toprule
Setting
& Avg.
& $>80\%$
& $>85\%$ \\
\midrule

w/o MCR--MSR
& 74.2$\pm$1.8
& 43.4$\pm$7.1
& 33.2$\pm$4.0 \\

w/o CGA
& 74.4$\pm$1.1
& 43.9$\pm$3.4
& 33.2$\pm$4.0 \\

w/o CAM
& 74.9$\pm$1.2
& 44.4$\pm$3.6
& 34.6$\pm$4.7 \\

w/o DAM
& 74.6$\pm$2.1
& 45.9$\pm$6.1
& 29.3$\pm$4.9 \\

w/o RAR
& 67.8$\pm$1.9
& 21.0$\pm$1.2
& 15.1$\pm$1.8 \\

\rowcolor{OursBlue}
\textbf{Ours}
& \textbf{77.8$\pm$0.3}
& \textbf{54.6$\pm$3.7}
& \textbf{37.1$\pm$3.2} \\
\bottomrule
\end{tabularx}
\caption{Ablation results on ChEMBL2K. The last two columns report the
percentage of tasks whose AUROC exceeds the specified threshold.}
\label{tab:ablation_study}
\end{table}

We evaluate the contributions of MCR--MSR disentanglement, CGA, CAM, DAM, and
RAR on ChEMBL2K. Table~\ref{tab:ablation_study} shows that removing MCR--MSR
disentanglement or CGA reduces the average AUROC by 3.6 and 3.4 percentage
points, respectively, supporting the roles of modality separation and consensus
alignment. Their comparable degradation indicates that isolating
modality-specific variation and consolidating cross-modal responses are both
necessary for constructing the PMRD space. Removing CAM lowers the AUROC to
74.9\%, indicating that local stability prevents the learned response factors
from relying excessively on incidental features.
Without DAM, the average AUROC decreases to 74.6\%, and the proportion of tasks
above 85\% drops by 7.8 percentage points. This larger decline at the stricter
threshold suggests that dynamic loss reweighting is particularly important for
maintaining strong performance on reliably predicted properties. RAR has the
largest observed effect: its removal reduces the average AUROC from 77.8\% to
67.8\% and decreases the proportions above 80\% and 85\% by 33.6 and 22.0
percentage points. 
\subsection{Representation Space Analysis}
\label{sec:representation_analysis}

Finally, we qualitatively examine whether the PMRD-center representation
contains biologically or assay-level coherent local structure. We first cluster
the embeddings in the original high-dimensional space and then conduct label
enrichment analysis for each cluster. The resulting cluster annotations are
visualized using t-SNE.

As shown in Figure~\ref{fig:mechanism_enriched_tsne}, several local regions are
associated with coherent enriched themes.
On ChEMBL2K, the PMRD-center representations form several locally coherent
regions associated with CNS receptor modulation~(CNS-R), carbonic anhydrase
inhibition~(CAI), inflammatory enzyme inhibition~(IEI), and kinase
signaling~(KS). Samples with weak enrichment~(WE) are more dispersed and partly
overlap with other regions, suggesting that drugs without a dominant
mechanistic theme are less clearly organized. In contrast, the enriched
clusters exhibit relatively compact local neighborhoods, indicating that drugs
sharing related biological responses tend to receive similar PMRD-center
representations despite differences in their individual modalities.
Broad6K exhibits a similar pattern at the assay level. Samples enriched for the
220\_692 biochemical assay~(B220), AF4-AF9 biochemical assay~(AF4-AF9),
\textit{C. neoformans} assay~(CNEO), Dengue viral assay~(DENV), HDL receptor
cell assay~(HDL-R), and LGR2 cell assay~(LGR2) occupy distinguishable local
regions. In particular, several assay groups form compact neighborhoods,
whereas others remain partially connected, reflecting both assay-specific
signals and response patterns shared across assays.
These observations support the PMRD design: CRM reduces modality-specific
interference, while CGA and SGO preserve cross-modal responses that are stable
and discriminative. The resulting PMRD center exhibits biologically meaningful
local structure, although t-SNE provides only qualitative evidence.

\section{Conclusion}

In this paper, we present a multimodal framework PMRD for mechanism-aware zero-shot drug property prediction. By separating mechanism-consistent factors from modality-specific information, PMRD constructs a consensus response domain and prioritizes signals that are cross-modally consistent, locally stable, and supportive of discriminative retrieval. Results on ChEMBL2K and Broad6K indicate improved predictive performance and more biologically coherent drug neighborhoods, while hard-negative analysis suggests fewer conflicts between structurally dissimilar but response-related compounds. Our future work will extend PMRD to additional profiling modalities and validate the retrieved mechanism relationships using external biological evidence.

\bibliography{aaai2027}

@inproceedings{robinson2021shortcut,
  title={Can contrastive learning avoid shortcut solutions?},
  author={Robinson, Joshua and Sun, Li and Yu, Ke and Batmanghelich, Kayhan and Jegelka, Stefanie and Sra, Suvrit},
  booktitle={Advances in Neural Information Processing Systems},
  volume={34},
  pages={4974--4986},
  year={2021}
}

@article{moshkov2023predicting,
  title={Predicting compound activity from phenotypic profiles and chemical structures},
  author={Moshkov, Nikita and Becker, Tim and Yang, Kevin and Horvath, Peter and Dancik, Vlado and Wagner, Bridget K. and Clemons, Paul A. and Singh, Shantanu and Carpenter, Anne E. and Caicedo, Juan C.},
  journal={Nature Communications},
  volume={14},
  number={1},
  pages={1967},
  year={2023}
}

@inproceedings{liu2025learning,
  title={Learning molecular representation in a cell},
  author={Liu, Gang and Seal, Srijit and Arevalo, John and Liang, Zhenwen and Carpenter, Anne E. and Jiang, Meng and Singh, Shantanu},
  booktitle={International Conference on Learning Representations},
  year={2025}
}

@inproceedings{ma2021conditional,
  title={Conditional contrastive learning for improving fairness in self-supervised learning},
  author={Ma, Martin Q. and Tsai, Yao-Hung Hubert and Liang, Paul Pu and Zhao, Han and Zhang, Kun and Salakhutdinov, Ruslan and Morency, Louis-Philippe},
  booktitle={International Conference on Learning Representations},
  year={2022}
}

@article{gaulton2012chembl,
  title={{ChEMBL}: A large-scale bioactivity database for drug discovery},
  author={Gaulton, Anna and Bellis, Louisa J. and Bento, A. Patricia and Chambers, Jon and Davies, Mark and Hersey, Anne and Light, Yvonne and McGlinchey, Shaun and Michalovich, David and Al-Lazikani, Bissan and Overington, John P.},
  journal={Nucleic Acids Research},
  volume={40},
  number={D1},
  pages={D1100--D1107},
  year={2012}
}

@inproceedings{rao2025miner,
  title={Multi-modal contrastive learning with negative sampling calibration for phenotypic drug discovery},
  author={Rao, Jiahua and Lin, Hanjing and Chen, Leyu and Xie, Jiancong and Zheng, Shuangjia and Yang, Yuedong},
  booktitle={Proceedings of the IEEE/CVF Conference on Computer Vision and Pattern Recognition},
  pages={30752--30762},
  year={2025}
}

@inproceedings{hu2020strategies,
  title={Strategies for Pre-training Graph Neural Networks},
  author={Hu, Weihua and Liu, Bowen and Gomes, Joseph and Zitnik, Marinka and Liang, Percy and Pande, Vijay and Leskovec, Jure},
  booktitle={International Conference on Learning Representations},
  year={2020}
}

@inproceedings{rong2020grover,
  title={Self-Supervised Graph Transformer on Large-Scale Molecular Data},
  author={Rong, Yu and Bian, Yatao and Xu, Tingyang and Xie, Weiyang and Wei, Ying and Huang, Wenbing and Huang, Junzhou},
  booktitle={Advances in Neural Information Processing Systems},
  volume={33},
  pages={12559--12571},
  year={2020}
}

@inproceedings{xu2021graphlog,
  title={Self-Supervised Graph-Level Representation Learning with Local and Global Structure},
  author={Xu, Minghao and Wang, Hang and Ni, Bingbing and Guo, Hongyu and Tang, Jian},
  booktitle={Proceedings of the 38th International Conference on Machine Learning},
  pages={11548--11558},
  year={2021}
}

@inproceedings{you2021joao,
  title={Graph Contrastive Learning Automated},
  author={You, Yuning and Chen, Tianlong and Shen, Yang and Wang, Zhangyang},
  booktitle={Proceedings of the 38th International Conference on Machine Learning},
  pages={12121--12132},
  year={2021}
}

@article{zheng2024crossmodal,
  title={Cross-modal Graph Contrastive Learning with Cellular Images},
  author={Zheng, Shuangjia and Rao, Jiahua and Zhang, Jixian and Zhou, Lianyu and Xie, Jiancong and Cohen, Ethan and Lu, Wei and Li, Chengtao and Yang, Yuedong},
  journal={Advanced Science},
  volume={11},
  number={32},
  pages={2404845},
  year={2024}
}

@inproceedings{radford2021learning,
  title={Learning Transferable Visual Models from Natural Language Supervision},
  author={Radford, Alec and Kim, Jong Wook and Hallacy, Chris and Ramesh, Aditya and Goh, Gabriel and Agarwal, Sandhini and Sastry, Girish and Askell, Amanda and Mishkin, Pamela and Clark, Jack and others},
  booktitle={International Conference on Machine Learning},
  pages={8748--8763},
  year={2021}
}

@inproceedings{wang2024infocore,
  title={Removing biases from molecular representations via information maximization},
  author={Wang, Chenyu and Gupta, Sharut and Uhler, Caroline and Jaakkola, Tommi S.},
  booktitle={International Conference on Learning Representations},
  year={2024}
}

@inproceedings{du2023unimodal,
  title={On uni-modal feature learning in supervised multi-modal learning},
  author={Du, Chenzhuang and Teng, Jiaye and Li, Tingle and Liu, Yichen and Yuan, Tianyuan and Wang, Yue and Yuan, Yang and Zhao, Hang},
  booktitle={Proceedings of the 40th International Conference on Machine Learning},
  pages={8632--8656},
  year={2023}
}

@article{jang2021moa,
  title={Predicting mechanism of action of novel compounds using compound structure and transcriptomic signature coembedding},
  author={Jang, Gwanghoon and Park, Sungjoon and Lee, Sanghoon and Kim, Sunkyu and Park, Sejeong and Kang, Jaewoo},
  journal={Bioinformatics},
  volume={37},
  number={Supplement\_1},
  pages={i376--i382},
  year={2021}
}

@article{wu2018moleculenet,
  title={MoleculeNet: A benchmark for molecular machine learning},
  author={Wu, Zhenqin and Ramsundar, Bharath and Feinberg, Evan N. and Gomes, Joseph and Geniesse, Caleb and Pappu, Aneesh S. and Leswing, Karl and Pande, Vijay},
  journal={Chemical Science},
  volume={9},
  number={2},
  pages={513--530},
  year={2018}
}

@article{yang2019dmpnn,
  title={Analyzing learned molecular representations for property prediction},
  author={Yang, Kevin and Swanson, Kyle and Jin, Wengong and Coley, Connor and Eiden, Philipp and Gao, Hua and Guzman-Perez, Angel and Hopper, Timothy and Kelley, Brian and Mathea, Miriam and others},
  journal={Journal of Chemical Information and Modeling},
  volume={59},
  number={8},
  pages={3370--3388},
  year={2019}
}

@article{wang2022molclr,
  title={Molecular contrastive learning of representations via graph neural networks},
  author={Wang, Yuyang and Wang, Jianren and Cao, Zhonglin and Barati Farimani, Amir},
  journal={Nature Machine Intelligence},
  volume={4},
  number={3},
  pages={279--287},
  year={2022}
}

@article{subramanian2017l1000,
  title={A next generation connectivity map: L1000 platform and the first 1,000,000 profiles},
  author={Subramanian, Aravind and Narayan, Rajiv and Corsello, Steven M. and Peck, David D. and Natoli, Ted E. and Lu, Xiaodong and Gould, Joshua and Davis, John F. and Tubelli, Andrew A. and Asiedu, Jacob K. and others},
  journal={Cell},
  volume={171},
  number={6},
  pages={1437--1452},
  year={2017}
}

@article{bray2017cellpainting,
  title={A dataset of images and morphological profiles of 30,000 small-molecule treatments using the Cell Painting assay},
  author={Bray, Mark-Anthony and Gustafsdottir, Sigrun M. and Rohban, Mohammad H. and Singh, Shantanu and Ljosa, Vebjorn and Sokolnicki, Katherine L. and Bittker, Joshua A. and Bodycombe, Nicole E. and Dancik, Vlado and Hasaka, Thomas P. and others},
  journal={GigaScience},
  volume={6},
  number={12},
  pages={giw014},
  year={2017}
}

@article{mcquin2018cellprofiler,
  title={CellProfiler 3.0: Next-generation image processing for biology},
  author={McQuin, Claire and Goodman, Allen and Chernyshev, Vasiliy and Kamentsky, Lee and Cimini, Beth A. and Karhohs, Kyle W. and Doan, Minh and Ding, Liya and Rafelski, Susanne M. and Thirstrup, Derek and others},
  journal={PLoS Biology},
  volume={16},
  number={7},
  pages={e2005970},
  year={2018}
}

@article{corsello2020prism,
  title={Discovering the anticancer potential of non-oncology drugs by systematic viability profiling},
  author={Corsello, Steven M. and Nagari, Rohith T. and Spangler, Ryan D. and Rossen, Jordan and Kocak, Mustafa and Bryan, Jordan G. and Humeidi, Ranad and Peck, David and Wu, Xiaoyun and Tang, Andrew A. and others},
  journal={Nature Cancer},
  volume={1},
  number={2},
  pages={235--248},
  year={2020}
}

@article{rogers2010extended,
  title={Extended-connectivity fingerprints},
  author={Rogers, David and Hahn, Mathew},
  journal={Journal of Chemical Information and Modeling},
  volume={50},
  number={5},
  pages={742--754},
  year={2010}
}

@inproceedings{gilmer2017neural,
  title={Neural message passing for quantum chemistry},
  author={Gilmer, Justin and Schoenholz, Samuel S. and Riley, Patrick F. and Vinyals, Oriol and Dahl, George E.},
  booktitle={Proceedings of the 34th International Conference on Machine Learning},
  pages={1263--1272},
  year={2017}
}

@article{xiong2020attentivefp,
  title={Pushing the boundaries of molecular representation for drug discovery with the graph attention mechanism},
  author={Xiong, Zhaoping and Wang, Dingyan and Liu, Xiaohong and Zhong, Feisheng and Wan, Xiaozhe and Li, Xutong and Li, Zhaojun and Luo, Xiaomin and Chen, Kaixian and Jiang, Hualiang and others},
  journal={Journal of Medicinal Chemistry},
  volume={63},
  number={16},
  pages={8749--8760},
  year={2020}
}

@inproceedings{you2020graphcl,
  title={Graph contrastive learning with augmentations},
  author={You, Yuning and Chen, Tianlong and Sui, Yongduo and Chen, Ting and Wang, Zhangyang and Shen, Yang},
  booktitle={Advances in Neural Information Processing Systems},
  year={2020}
}

@inproceedings{zhou2023unimol,
  title={Uni-Mol: A universal 3D molecular representation learning framework},
  author={Zhou, Gengmo and Gao, Zhifeng and Ding, Qiankun and Zheng, Hang and Xu, Hongteng and Wei, Zhewei and Zhang, Linfeng and Ke, Guolin},
  booktitle={International Conference on Learning Representations},
  year={2023}
}

@article{sanchez2023cloome,
  title   = {{CLOOME}: Contrastive Learning Unlocks Bioimaging Databases
             for Queries with Chemical Structures},
  author  = {Sanchez-Fernandez, Ana and Rumetshofer, Elisabeth and
             Hochreiter, Sepp and Klambauer, G{\"u}nter},
  journal = {Nature Communications},
  volume  = {14},
  number  = {1},
  pages   = {7339},
  year    = {2023},
  doi     = {10.1038/s41467-023-42328-w}
}

@article{zheng2024miga,
  title   = {Cross-Modal Graph Contrastive Learning with Cellular Images},
  author  = {Zheng, Shuangjia and Rao, Jiahua and Zhang, Jixian and
             Zhou, Lianyu and Xie, Jiancong and Cohen, Ethan and
             Lu, Wei and Li, Chengtao and Yang, Yuedong},
  journal = {Advanced Science},
  volume  = {11},
  number  = {32},
  pages   = {2404845},
  year    = {2024},
  doi     = {10.1002/advs.202404845}
}

@inproceedings{radford2021clip,
  title     = {Learning Transferable Visual Models from Natural Language Supervision},
  author    = {Radford, Alec and Kim, Jong Wook and Hallacy, Chris and
               Ramesh, Aditya and Goh, Gabriel and Agarwal, Sandhini and
               Sastry, Girish and Askell, Amanda and Mishkin, Pamela and
               Clark, Jack and Krueger, Gretchen and Sutskever, Ilya},
  booktitle = {Proceedings of the 38th International Conference on Machine Learning},
  series    = {Proceedings of Machine Learning Research},
  volume    = {139},
  pages     = {8748--8763},
  year      = {2021},
  publisher = {PMLR}
}

@inproceedings{stark2022infomax,
  title={3D Infomax improves GNNs for molecular property prediction},
  author={St{\"a}rk, Hannes and Beaini, Dominique and Corso, Gabriele and Tossou, Prudencio and Dallago, Christian and G{\"u}nnemann, Stephan and Li{\`o}, Pietro},
  booktitle={Proceedings of the 39th International Conference on Machine Learning},
  pages={20479--20502},
  year={2022}
}

@article{chandrasekaran2023jump,
  title={JUMP Cell Painting dataset: Morphological impact of 136,000 chemical and genetic perturbations},
  author={Chandrasekaran, Srinivas Niranj and others},
  journal={bioRxiv},
  year={2023}
}
\end{document}